\documentclass[11pt]{article}

\usepackage[preprint]{acl}

\usepackage{times}
\usepackage{latexsym}
\usepackage{helvet}
\usepackage{courier}
\usepackage{amsmath}
\usepackage{amsfonts}
\usepackage{booktabs}
\usepackage{graphicx}  
\usepackage{booktabs}  
\usepackage{multirow} 
\usepackage{comment}
\usepackage{placeins}
\usepackage[skins,breakable]{tcolorbox}
\newcommand{\tcite}[1]{{\scriptsize \cite{#1}}}
\newtcolorbox{promptbox}[1]{
  enhanced,
  breakable,
  colback=gray!4,
  colframe=black!60,
  title={#1},
  fonttitle=\bfseries,
  fontupper=\small,
  boxrule=0.4pt,
  arc=1pt,
  left=5pt,
  right=5pt,
  top=5pt,
  bottom=5pt
}
\usepackage[T1]{fontenc}

\usepackage[utf8]{inputenc}

\usepackage{microtype}

\usepackage{inconsolata}

\title{CoLa-ICD: A Knowledge-Enhanced Framework for Long-Tail Automated Medical Coding}

\author{
Yihang Cheng \thanks{Corresponding author.} , Veronica Liesaputra , Andrew Trotman \\
University of Otago \\
Dunedin, New Zealand \\
\texttt{yihang.cheng@postgrad.otago.ac.nz} \\
\texttt{veronica.liesaputra@otago.ac.nz} \\
\texttt{andrew.trotman@otago.ac.nz}
}

\begin{document}
\maketitle
\begin{abstract}
Automatic medical coding assigns ICD codes to clinical notes, but it remains challenging due to long documents, imbalanced label distributions, and diverse terms. These challenges are especially severe for rare codes, which have limited training instances and are easily confused with semantically similar labels. We introduce \textbf{CoLa-ICD}, a knowledge-enhanced framework for long-tail prediction. CoLa-ICD enriches ICD labels with external terms, models dependencies among related codes, and learns stronger alignment between label semantics and clinical evidence for long-tail prediction. Experiments show that CoLa-ICD improves long-tail prediction with larger gains in larger and sparser label spaces and achieves state-of-the-art performance in AUC, F1, and P@$k$.  Our code is available at \url{https://github.com/youwillbethebest/Cola-ICD}.
\end{abstract}

\section{Introduction}

The International Classification of Diseases (ICD)\footnote{\scriptsize https://www.who.int/standards/classifications/classificationof-diseases} creates a standard system for medical billing, epidemiology, and clinical analytics \cite{ellis_diagnostic_2020,feinstein_preparing_2023}. Automated medical coding (AMC) aims to assign multiple ICD codes to discharge summaries, posing a challenging multi-label classification problem \cite{stanfill_systematic_2010}.
\begin{figure}[htbp]       
  \centering
  \includegraphics[width=0.9\linewidth]{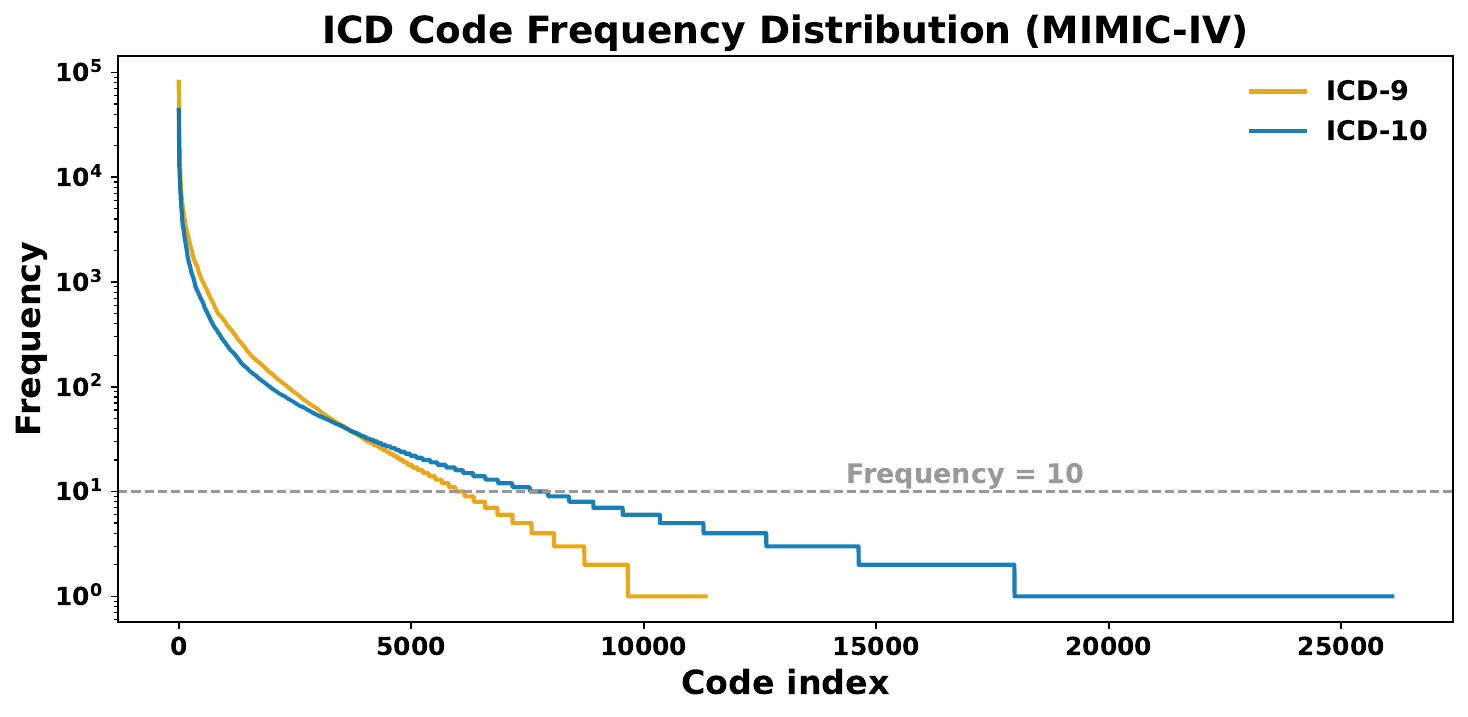}
  \caption{Frequency distribution of ICD codes in MIMIC-IV: Most codes occur fewer than 10 times, illustrating the long-tailed nature of the ICD coding task.}
  \label{fig1}
\end{figure}
AMC remains difficult for two reasons \cite{edin_automated_2023,huang_plm-icd_2022}. First, ICD codes follow a highly long-tailed distribution (Figure~\ref{fig1}), with more than half appearing fewer than 10 times, making long-tail codes extremely hard to learn. Predicting long-tail ICD codes is vital for capturing rare diseases and precise clinical details that common codes overlook, ensuring equitable care and accurate medical billing. Second, clinical notes are long (1,500--2,000 words), but each ICD code is often supported by only a few localized, semantically dense cues, such as a diagnosis mention, procedure phrase, medication, or abnormal finding. Figure~\ref{fig2} illustrates another challenge: clinical notes may express the same concept using wording that differs from the official ICD description. For example, ``History of tobacco use'' could be written as ``former smoker'' or ``quit tobacco''. This mismatch is particularly detrimental for long-tail codes, whose semantics must be inferred from a limited number of training instances.

Recent research \cite{ji_unified_2024,li_deep_2025} leverages external medical knowledge to improve automated coding. Existing approaches fall into two groups: Large Language Models (LLMs) and static, task-specific methods. LLMs (e.g., ChatGPT \cite{achiam_gpt-4_2024} and Gemini \cite{anil_gemini_2025}) provide strong medical knowledge and flexible generation. However, directly applying these generative models often leads to overfitting on high-frequency codes and generating hallucinated or incorrect codes \cite{falis_can_2024,soroush_large_2024,hou_enhancing_2025}. Task-specific methods use fixed synonyms \cite{yuan_code_2022,gomes_accurate_2024} or rigid hierarchical graphs \cite{yang_knowledge_2022,zhang_general_2025} to ground predictions. They fail to capture the diverse abbreviations, standard terms, and informal expressions found in real-world clinical notes. Furthermore, relying solely on static hierarchical distances offers limited improvement for long-tail labels, leaving models prone to confusing rare codes with similar frequent ones \cite{zhou_automatic_2021}.

\begin{figure}[htbp]      
  \centering
  \includegraphics[width=0.9\linewidth]{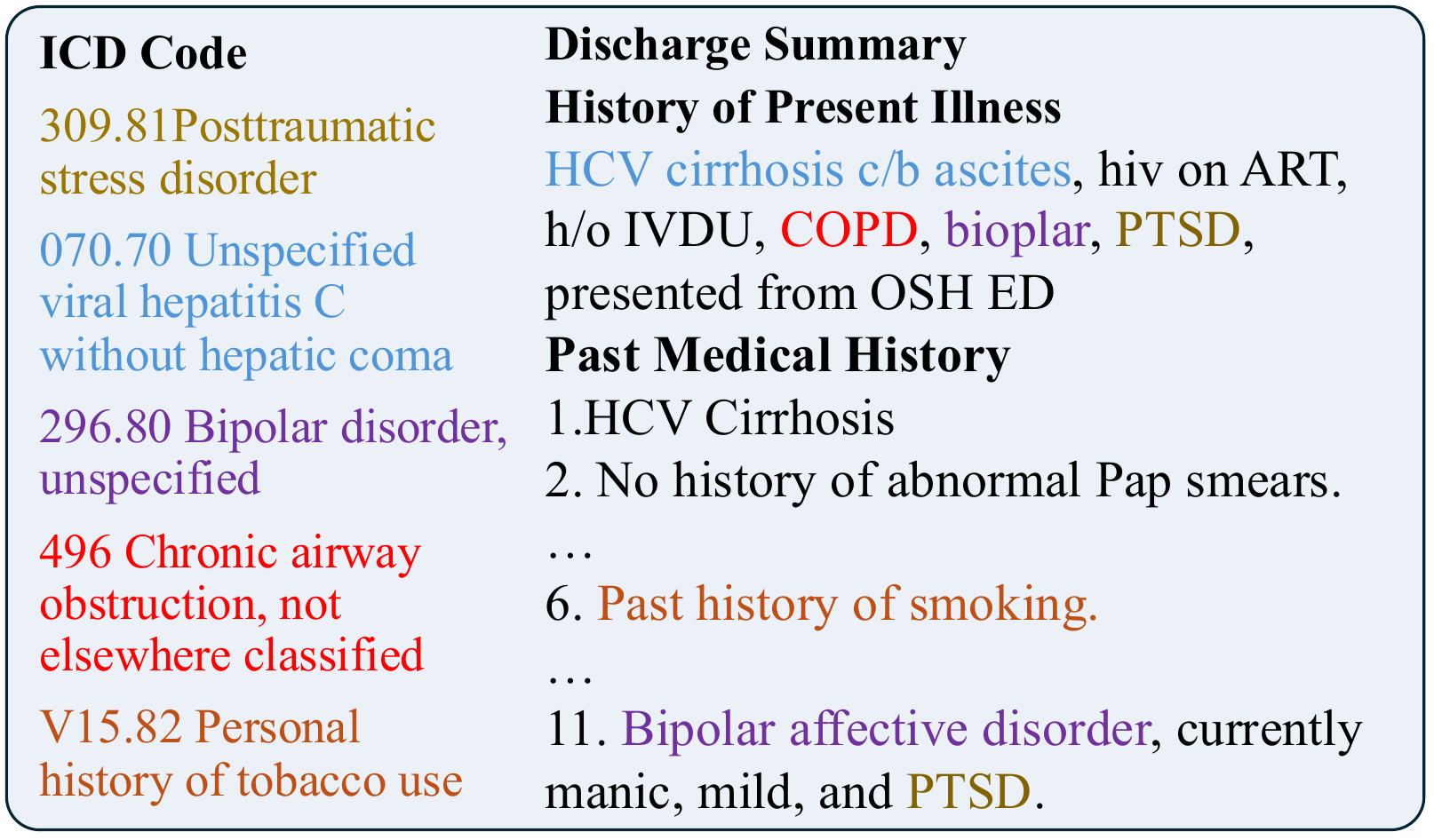}
  \caption{A discharge summary with ICD-9 codes. Colored spans show the short evidence for each label.}
  \label{fig2}
\end{figure}

To address these limitations, CoLa-ICD combines LLM-generated terms, co-occurrence relations, and label-aware contrastive learning to improve evidence-label alignment for long-tail codes.

Our contributions are summarized as follows:
\begin{itemize}
     \item We propose a knowledge-enhanced framework that enriches label representations with LLM-generated terms, integrates a co-occurrence GNN to infuse code relations, and uses multi-head term attention to retrieve supporting evidence from clinical text. This is particularly beneficial for long-tail codes.
     \item  We introduce a label-aware contrastive learning module that aligns clinical evidence with enriched label representations. This mechanism enforces evidence–label consistency and mitigates the bias toward high-frequency codes. This is critical for identifying long-tail and semantically similar labels.
     \item We achieve state-of-the-art performance on ICD-9 and ICD-10 benchmarks, with robust improvements on long-tail codes.
\end{itemize}

\section{Related Work}
\subsection{Automatic Medical Coding}
Automatic medical coding aims to assign ICD codes to discharge summaries with deep learning models. Early methods  \cite{mullenbach_explainable_2018,chen_multi-channel_2019, Li2020ICDNetwork,vu_label_2020,ji_dilated_2020} use convolutional or recurrent neural networks. These approaches created strong and clear baselines.

Instead of pooling a document into a single embedding, label-wise attention measures the relevance of each token to a specific label \cite{mullenbach_explainable_2018}. LAAT \cite{vu_label_2020} deepens label attention with multi-level layers, leading to better alignment between label semantics and evidence tokens in the clinical text. Follow-up works \cite{biswas_transicd_2021,wu_jan_2022,edin_automated_2023} replace label-wise attention with cross-attention, enabling richer interactions between code and text.

\subsection{Pretrained Language Models}
Pretrained language models (PLMs), such as BioBERT \cite{lee_biobert_2020} have been widely adopted for AMC \cite{huang_plm-icd_2022,yang_knowledge_2022}.  By pretraining on large-scale biomedical corpora (e.g., PubMed, MIMIC, UMLS), these models acquire domain-specific knowledge, improving the semantic representations for discharge summaries. A parallel line of work has focused on handling long clinical notes. Segment pooling \cite{zhang_bert-xml_2020,edin_unsupervised_2024} and chunk-based strategies \cite{liu_automated_2023} solve the max token limitation of BERT encoders by dividing text into segments. However, while PLMs improve representation quality, they still struggle with aligning clinical evidence with ICD codes. More recently, LLMs have been explored for AMC through code generation. Although their broad biomedical knowledge and few-shot abilities make them attractive for clinical coding, recent studies show that LLMs remain unreliable for exact ICD assignment, especially on challenging MIMIC-IV cases \cite{soroush_large_2024,mustafa_large_2025}. Some studies \cite{hou_enhancing_2025,motzfeldt-etal-2025-code} indicate that LLM can improve performance through task-specific adaptation, retrieval, or workflow constraints.

\subsection{Knowledge Injection}

With the development of representation learning \cite{zhang2026faithful}. Recent studies have improved AMC performance by enriching ICD labels with external knowledge in two ways.

For structure-based methods, MSATT-KG \cite{xie_ehr_2019} uses Graph Convolutional Networks to capture hierarchical and semantic relationships across related codes. KEPTLongformer \cite{yang_knowledge_2022} improves representation via contrastive learning on knowledge graphs in the pretraining stage. For text-based methods, MSMN \cite{yuan_code_2022} expands official ICD descriptions with multiple synonyms, each used as a unique query to match evidence in clinical texts. GKI-ICD \cite{zhang_general_2025} flattens the hierarchical relations into textual label definitions, treating structure as semantic context. PLM-LLM \cite{wu_contrastive_2025} used GPT-4 to generate code synonyms, which are utilized as positive samples via contrastive learning. Different from injecting external knowledge, CoRelation \cite{luo_corelation_2024} focuses on mining internal structural knowledge. It captures label correlations directly from clinical notes in order to enhance prediction consistency. However, the scarcity of training data hinders the model from learning stable and reliable rare code structures.

Although existing knowledge-injection methods can improve AMC performance by enriching label semantics, they only partially align with practical coding practices and often fail to address long-tail prediction. In particular, they often address either semantic sparsity in code descriptions or structural sparsity in code dependencies, but seldom model both within the same evidence-label matching process. This gap limits their ability to ground evidence, especially for long-tail cases and similar codes.

\subsection{Contrastive Learning}

Contrastive learning has been explored to enhance semantic representations for ICD labels. HyperCore \cite{cao_hypercore_2020} embeds documents and codes into a shared hyperbolic space to maintain hierarchical consistency between high-level codes and their subcategories. KEPTLongformer \cite{yang_knowledge_2022} uses contrastive learning that leverages the hierarchical structure of the ICD code. It treats synonyms as positive matches and hierarchical codes as negative matches to improve long-tail performance.  \citeauthor{li_towards_2023} introduce a section-level contrastive learning method. They build groups of four samples and extract positive pairs from different sections of clinical notes. PLM-LLM \cite{wu_contrastive_2025} creates positives using LLM-generated paraphrases of the official ICD description.

These approaches enhance semantic representations and outperform non-contrastive baselines. However, most contrastive methods look at the text and label as a whole. They fail to notice short, specific pieces of evidence. These small details are necessary for accurate ICD coding \cite{mullenbach_explainable_2018,vu_label_2020}.
\section{Problem Formulation}
Assigning ICD codes to clinical documents can be framed as a multi-label text classification problem. Given a clinical document $X$, the goal is to automatically predict all relevant ICD codes from a large set of possible codes. Let
\(
\mathcal{C} = \{ c_1, c_2, \dots, c_{N} \}
\)
represent the set of all candidate ICD codes, where \( N \) is the total number of codes. Each document is associated with a binary vector
\begin{equation}
    Y = \{ y_i \in \{0,1\} \mid i \in \{1,\dots,N\} \, \}
\end{equation}
where $y_i = 1$ if code $c_i$ applies to the document, and $y_i = 0$ otherwise. The AMC task is challenging due to the large label space and the highly imbalanced, long-tailed distribution of code frequencies.

\section{Methods}
\begin{figure*}[t]      
  \centering
  \includegraphics[width=0.85\linewidth]{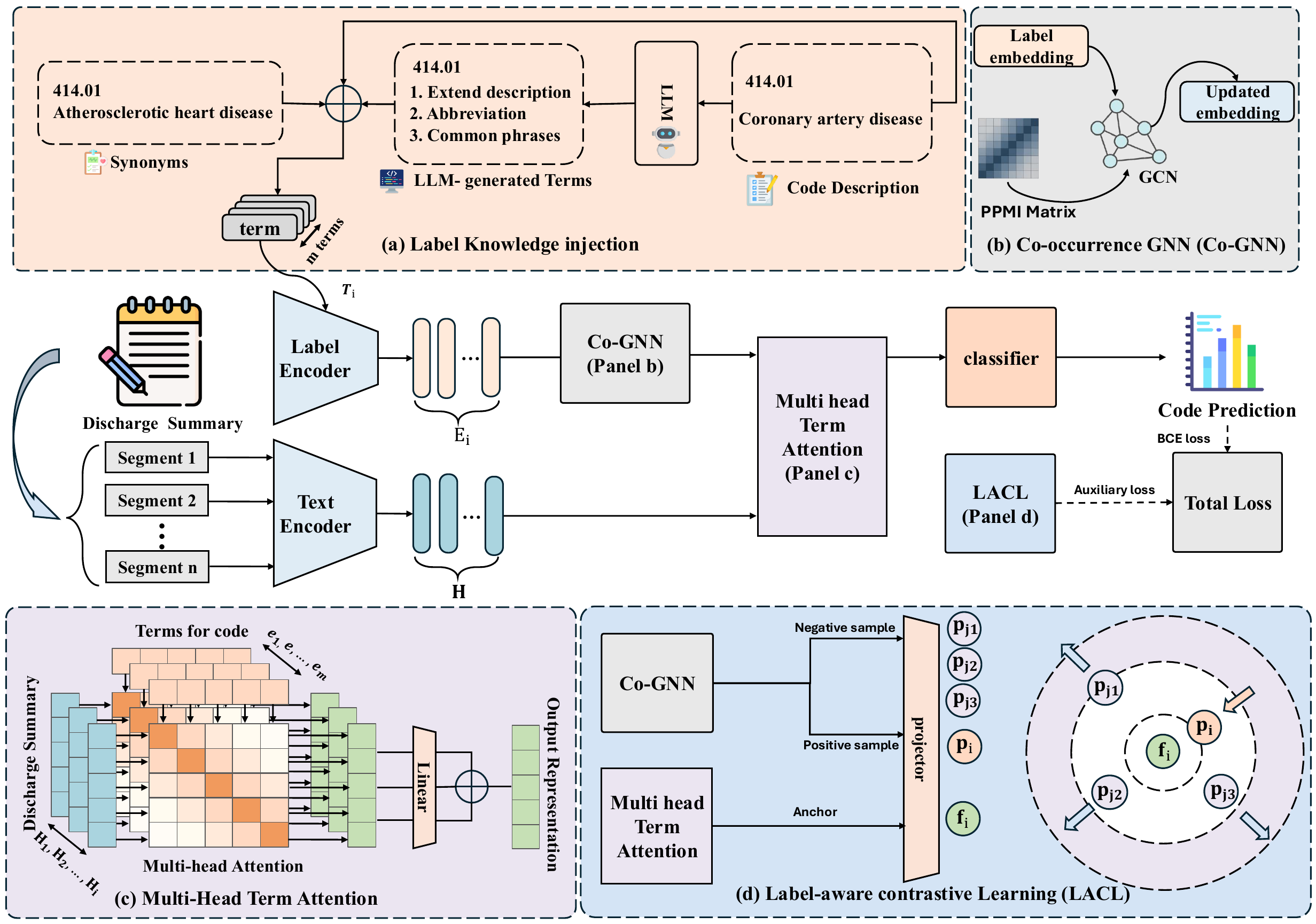}
  \caption{Overall architecture of CoLa-ICD. \textbf{(a) Label Knowledge Injection} constructs the label knowledge set $T_i$ for each ICD code. The encoded terms are passed to the \textbf{(b) Co-occurrence GNN}, which produces knowledge-enhanced label representations $p_i$. The GNN-updated label-term representations are then used by \textbf{(c) Multi-Head Term Attention} over contextual token representations to extract an evidence representation $f_i$ for classification. During training, \textbf{(d) Label-Aware Contrastive Learning} receives $f_i$ from \textbf{(c) Multi-Head Term Attention} as anchor and $p_i$ from the \textbf{(b) Co-occurrence GNN}, and optimizes an auxiliary contrastive loss that pulls $f_i$ toward its positive label embedding while pushing it away from negative label embeddings.}
  \label{fig3}
\end{figure*}
Our framework, illustrated in Figure~\ref{fig3} improves AMC from two perspectives. It strengthens label embeddings by leveraging external knowledge and code relations found in clinical documents.
Furthermore, it improves semantic alignment between clinical evidence and ICD codes via contrastive learning.
 
\subsection{Segment Pooling}
We follow a segment method inspired by PLM-ICD \cite{huang_plm-icd_2022}. Given a discharge summary with $T$ tokens, we split it into $N_{\mathrm{seg}}$ segments. Each segment is encoded by SapBERT \cite{liu_self-alignment_2021}. SapBERT is pre-trained to align synonymous medical entities in UMLS, which is well-suited to handling term variation between clinical notes and ICD codes.
We then concatenate the segment-level hidden states along the token dimension:
\begin{equation} 
H=\operatorname{concat}(H_1,\ldots,H_{N_{\mathrm{seg}}})\in\mathbb{R}^{T\times d}
\end{equation}
This strategy preserves all tokens in the discharge summary without truncation and enables the model to attend to evidence scattered across long documents. 
\subsection{Label Knowledge Injection}
\label{sec_know_inject}

We enrich each ICD code with multiple knowledge sources to bridge the gap between sparse training data and diverse clinical expressions. Specifically, we use the Unified Medical Language System (UMLS, 2024AB release) \cite{bodenreider_unified_2004} to obtain the official ICD description $\ell_i$ and synonyms $s_i$. In addition, we query a large language model (Gemini 2.5 Flash \cite{comanici_gemini_2025}) to generate augmented terms for each code, including abbreviations, extended definitions, and common clinical phrases. We select the term-generation strategy using the MIMIC-III validation split and keep it fixed across all datasets. The prompts and selection procedure are provided in Appendix~\ref{prompts for term} and \ref{app:llm_terms_selection}. For each code $c_i$, we build a fixed-length term list $T_i$ of size $m$ containing the description, LLM-generated expressions (abbreviations, extended descriptions, common phrases) and synonyms.
\begin{equation} 
T_i=[t^{i}_{1},\dots,t^{i}_{m}]
\end{equation}

Then, for each code $c_i$, we encode the term list into $E_i\in \mathbb{R}^{m\times d}$ and flatten all code-term embeddings into $E\in\mathbb{R}^{(N m)\times d}$ for the co-occurrence graph module.
\subsection{Co-occurrence Graph Neural Network}
To capture the pattern of ICD codes in real clinical notes, we use a co-occurrence GNN module. Based on a multi-hot label matrix $Y$ over training set, we compute a co-occurrence matrix $R=Y^\top Y$, where $R_{ij}$ denotes how frequently codes $c_i$ and $c_j$ co-occur in the training set.  Raw co-occurrence counts are often dominated by common codes that appear frequently by chance. We transform $R$ into a PPMI matrix $A\in\mathbb{R}^{N\times N}$ to down-weight such frequency-driven associations and highlight meaningful dependencies.

\begin{equation}
\mathrm{PPMI}(i,j)
= \max\!\left(
\log \frac{\Pr(i,j)}{\Pr(i)\Pr(j)},\, 0
\right)
\end{equation}
where $\Pr(i)$ and $\Pr(j)$ denote the marginal probabilities of 
codes $c_i$ and $c_j$, and $\Pr(i,j)$ denotes their joint probability. For each code node, we retain the top 10 edges with the highest PPMI values, based on validation performance. To convert the PPMI matrix into an adjacency matrix suitable for GNN, we expand $A$ into a term-level matrix $A\in\mathbb{R}^{(N m)\times(N m)}$. For each head $h$, edges are added only between the corresponding $h^{\text{th}}$ terms of co-occurring codes. We then add self-loops:
\begin{equation}
    \hat{A} = D^{-\frac{1}{2}}(A+I)\,D^{-\frac{1}{2}}
\end{equation}
where $I$ is the identity matrix and $D$ denotes the degree matrix. Then, we use term embedding $E$ as input for a two-layer GCN.
\begin{equation}
    Z = \hat{A}\,\sigma\!\left(\hat{A} E W_{0}\right) W_{1}
\end{equation}
where $W_0$ and $W_1$ are trainable weights. The updated embedding $Z$ is subsequently
fed into the Multi-head term attention module.

\subsection{Multi-head Term Attention}
\label{sec:term_attention}

To extract evidence with multiple semantic views, we adopt a multi-head mechanism. The key idea is to treat $m$ terms associated with code $c_i$ as distinct queries to extract relevant information from clinical notes. Given the document representation $H\in\mathbb{R}^{T\times d}$, we split it into $m$ heads based on the number of terms. Each head corresponds to a subspace representation  $H^{(h)}\in \mathbb{R}^{T \times \frac{d}{m}}$.

Let $E_i = \{e_1^i, \ldots, e_m^i\}$ denote the embeddings of the $m$ terms for code $c_i$, where $e_h^i \in \mathbb{R}^d$. The projection $W_Q e_h^i \in \mathbb{R}^{d/m}$ matches the dimensionality of the corresponding head representation $H^{(h)}$.
For the $h^{\text{th}}$ head, which corresponds to $e^{i}_{h}$, we compute the attention scores over the document tokens.
\begin{equation}
\alpha_i^{(h)} = \operatorname{softmax}\!\big(W_Q e^{i}_{h} \cdot \tanh(W_H H^{(h)})^\top\big)
\end{equation}

The evidence representation for head $h$ is then obtained by aggregating the subspace information
\begin{equation}
v_i^{(h)} = \alpha_i^{(h)} H^{(h)} \in \mathbb{R}^{ d/m}
\end{equation}
We then concatenate evidence from all heads to form the final evidence representation $f_i$.

\begin{equation}
f_i=\text{concat}(v_i^1,v_i^2,...,v_i^m) \in \mathbb{R}^d
\end{equation}

Finally, $f_i$ is passed through a multi-layer perceptron (MLP) followed by a sigmoid activation to produce the prediction score for code $c_i$:
\begin{equation}
\hat{y}_i = \sigma\!\big(\mathrm{MLP}(f_i)\big).
\end{equation}
and the same evidence embedding $f_i$ is used as the anchor representation in the label-aware contrastive learning.

\subsection{Label-Aware Contrastive Learning}
\label{sec:contrastive}
To align clinical evidence with the semantics of ICD labels, we introduce a Label-Aware Contrastive Learning (LACL) module. Given a discharge summary \(X\) and a candidate label \(c_i\), the multi-head term attention module extracts an evidence embedding \(f_i\) that serves as the anchor. We obtain a knowledge-enhanced embedding \(p_i\) for each code based on the GCN-updated representations derived from the label knowledge $T_i$.

For each pair $(f_i, p_i)$, we construct a contrastive candidate set $P$, consisting of all positive labels and $K$ negative labels. We first compute the cosine similarities between the anchor $f_i$ and all candidate label embeddings. 
To identify hard negatives $\mathcal{N}_i^{\text{hard}}$, we exclude positive label $p_i$ and select the top $\lfloor \rho K \rfloor$ with the highest similarity scores. These codes are semantically close to the anchor, but they are incorrect. The remaining $K - \lfloor \rho K \rfloor$ negatives are uniformly sampled. This hard-negative mining strategy forces the model to distinguish evidence representations for semantically similar codes.

The LACL loss is:
\begin{equation}
\label{lacl}
\mathcal{L}_{LACL} =
- \log 
\frac{\exp\!\big(\mathrm{sim}(f_i,p_i)/\tau\big)}
{\sum_{j \in P} \exp\!\big(\mathrm{sim}(f_i,p_j)/\tau\big)}
\end{equation}
where $\mathrm{sim}(\cdot,\cdot)$ denotes cosine similarity and $\tau$ is a temperature hyperparameter. This evidence–label contrast regularizes the attention-extracted evidence $f_i$ to be consistent with the corresponding knowledge-enhanced label embedding $p_i$, while pushing it away from other label semantics.

\begin{table*}[!t]
\setlength{\tabcolsep}{4pt}
\renewcommand{\arraystretch}{0.81}
\scriptsize
\centering
\resizebox{0.8\textwidth}{!}{
\begin{tabular}{lcccccc}
\toprule
\textbf{Dataset}
&\textbf{Code Freq.}&\textbf{Code Num.} & \textbf{PLM-CA}&\textbf{MSMN} & \textbf{GKI-ICD}&\textbf{CoLa-ICD (Ours)} \\
\midrule
\multirow{5}{*}{MIMIC-III ICD-9}
&$>$500 &(283) & 0.684& 0.658& \underline{0.687}&\textbf{0.693}\textsuperscript{*} \\
&101--500 &(737) & 0.508& 0.461& \underline{0.509}&\textbf{0.513}\textsuperscript{*} \\
&51--100 &(513) & 0.413&0.332 & \underline{0.420}&\textbf{0.427}\textsuperscript{*} \\
&11--50 &(1823)  & 0.293&0.227 & \underline{0.322}&\textbf{0.338} \textsuperscript{*}\\
&1--10 &(5336)   & 0.029&0.070 & \underline{0.132}&\textbf{0.194}\textsuperscript{*} \\
\midrule
\multirow{5}{*}{$\text{MIMIC-IV ICD-10}$}
&$>$500 &(480) & 0.598& 0.618& \underline{0.632}&\textbf{0.659}\textsuperscript{*}  \\
&101--500& (1067) & 0.425& 0.451& \underline{0.466}&\textbf{0.498}\textsuperscript{*} \\
&51--100 &(960) & 0.273&0.354 & \underline{0.398}&\textbf{0.426}\textsuperscript{*} \\
&11--50 &(3906)  & 0.153&0.272 & \underline{0.303}&\textbf{0.359} \textsuperscript{*}\\
&1--10 &(19532)   & 0.023&0.095 & \underline{0.127}&\textbf{0.204} \textsuperscript{*} \\
\bottomrule
\end{tabular}}
\caption{
Across both MIMIC-III and IV, our CoLa-ICD significantly outperforms existing approaches in frequency-bucket Micro-F1.
\textsuperscript{*} denotes $p<0.05$ after Holm-Bonferroni correction under the Wilcoxon signed-rank test.}
\label{tab:code_freq}
\end{table*}

\subsection{Training Objective}
Our model performs multi-label classification with one sigmoid output per ICD code. For a document with ground-truth label vector $y \in \{0,1\}^N$ and predicted probabilities $\hat{y} \in [0,1]^N$, we use the binary cross-entropy loss:
\begin{equation}
\mathcal{L}_{\text{BCE}} = -\frac{1}{N} \sum_{i=1}^{N} \Big[ y_i \log(\hat{y}_i) + (1 - y_i)\log(1 - \hat{y}_i) \Big],
\end{equation}
where $\hat{y}_i$ is the predicted probability that code $c_i$ applies to the document and $y_i$ is its ground-truth label.

To further enforce semantic consistency between textual evidence and enriched label representations, we add the label-aware contrastive learning loss defined in Section~\ref{sec:contrastive}. The overall training objective is
\begin{equation}
\mathcal{L} = \mathcal{L}_{\text{BCE}} + \lambda \,\mathcal{L}_{\text{LACL}},
\end{equation}
where $\lambda$ is the weight of the contrastive learning. In our experiments, we set $\lambda = 0.05$ based on performance on validation set. This value was selected to balance the magnitude differences between the two loss terms. 

\section{Experiments}
\subsection{Datasets}
We focus on the commonly used MIMIC-III ICD-9 \cite{johnson_mimic-iii_2016} and MIMIC-IV ICD-10 \cite{johnson_mimic-iv_2023} datasets, and follow \citeauthor{mullenbach_explainable_2018,edin_automated_2023}'s protocol to obtain all datasets (dataset statistics are provided in Appendix~\ref{app:dataset_stats}). MIMIC-IV ICD-10 contains a larger and sparser code space than MIMIC-III ICD-9, intensifying the long-tail challenge. We report MIMIC-III Top50 and MIMIC-IV ICD-9 results in Appendix~\ref{app:supp_results} for comparability with prior work.

\subsection{Metrics}
We report Micro-F1 and Macro-F1 as classification metrics, Micro-AUC and Macro-AUC as discrimination metrics, and Precision@$k$ as ranking metrics. Following prior work on ICD coding \cite{mullenbach_explainable_2018}, we set $k=8$ and $15$ for evaluation on the full MIMIC-III dataset. For the MIMIC-IV ICD-10, we use $k=8$.
\subsection{Implementation Details}
We implement CoLa-ICD using PyTorch \cite{paszke_pytorch_2019} on a single NVIDIA A100 80GB GPU. We optimize the model with AdamW, using an initial learning rate of $2\times10^{-5}$, a batch size of 6, cosine annealing with 2,000 warm-up steps, and early stopping over at most 20 epochs. For LACL, we set the hard-negative fraction $\rho=0.3$, temperature $\tau=0.1$, and number of negative samples $K=128$ for the main full-label experiments. The prediction threshold is selected on the validation set by maximizing Micro-F1. All experiments are repeated with ten random seeds, and we report the average performance. Additional hyperparameter details are provided in Appendix~\ref{app:hyperparams}.

We compare CoLa-ICD against two categories of baselines. The first category includes models without knowledge injection, including CAML \cite{mullenbach_explainable_2018}, PLM-ICD \cite{huang_plm-icd_2022}, and PLM-CA \cite{edin_unsupervised_2024}. The second category includes knowledge-enhanced methods, including MSMN \cite{yuan_code_2022}, KEPTLongformer \cite{yang_knowledge_2022}, CoRelation \cite{luo_corelation_2024}, and GKI-ICD \cite{zhang_general_2025}.

\begin{table*}[t]

\centering
\small
\resizebox{0.9\textwidth}{!}{
\begin{tabular}{lcccccc ccccc}
\toprule
\multirow{3}{*}{\textbf{Models}} &
\multicolumn{6}{c}{\textbf{MIMIC-III ICD-9}} &
\multicolumn{5}{c}{\textbf{MIMIC-IV ICD-10}} \\
\cmidrule(lr){2-7}\cmidrule(lr){8-12}
& \multicolumn{2}{c}{AUC} & \multicolumn{2}{c}{F1} & \multicolumn{2}{c}{P@K}
& \multicolumn{2}{c}{AUC} & \multicolumn{2}{c}{F1} & P@8 \\
\cmidrule(lr){2-3} \cmidrule(lr){4-5} \cmidrule(lr){6-7} \cmidrule(lr){8-9}\cmidrule(lr){10-11}
& Macro & Micro & Macro & Micro & P@8 & P@15
& Macro & Micro & Macro & Micro &  \\
\midrule
CAML \tcite{mullenbach_explainable_2018}    & 0.895 & 0.986 & 0.088 & 0.539 & 0.709 & 0.561 & 0.899 & 0.988 & 0.046 & 0.527 & 0.644 \\
MSMN \tcite{yuan_code_2022}                 & 0.950 & 0.992 & 0.103 & 0.584 & 0.752 & 0.599 & 0.971 & 0.996 & 0.054 & 0.559 & 0.677 \\
KEPTLongformer \tcite{yang_knowledge_2022}  &   --  &   --  & 0.118 & 0.599 & 0.771 & 0.615 & -- & -- & -- & -- & -- \\
PLM-ICD \tcite{huang_plm-icd_2022}          & 0.926 & 0.989 & 0.104 & 0.598 & 0.771 & 0.613 & 0.919 & 0.990 & 0.049 & 0.567 & 0.695 \\
PLM-CA \tcite{edin_unsupervised_2024}       &0.916&0.989&0.103&0.599&0.772&0.616&0.920&0.990&0.052&0.570&0.699 \\
CoRelation \tcite{luo_corelation_2024}      &0.952&0.992&0.102&0.591&0.762&0.607&\underline{0.972}&0.996&0.063&0.578&0.700\\
GKI-ICD \tcite{zhang_general_2025} &\underline{0.962}&\underline{0.993}&\underline{0.123}&\underline{0.612}&\underline{0.777}&\underline{0.624}&0.971&\textbf{0.997}&\underline{0.069}&\underline{0.579}&\underline{0.702} \\
\textbf{CoLa-ICD (Ours)} &\textbf{0.969}\textsuperscript{*}&\textbf{0.998}\textsuperscript{*}&\textbf{0.135}\textsuperscript{*}& \textbf{0.621}\textsuperscript{*}&\textbf{0.781}\textsuperscript{*}&\textbf{0.633}\textsuperscript{*}&\textbf{0.977}\textsuperscript{*}&\textbf{0.997}&\textbf{0.108}\textsuperscript{*}&\textbf{0.600}\textsuperscript{*}&\textbf{0.723}\textsuperscript{*}\\
\bottomrule
\end{tabular}}
\caption{Results on MIMIC-III 
and MIMIC-IV
. The best scores among all models for each metric are highlighted in bold and underlined numbers denote the second-best. \textsuperscript{*} denotes $p<0.05$ after Holm-Bonferroni correction under the Wilcoxon signed-rank test.}
\label{tab:results}
\end{table*}

\subsection{Results}
\subsubsection{Performance on Long-tailed Codes}
Because CoLa-ICD is designed to improve coding under label sparsity, we first evaluate Micro-F1 across five training-frequency buckets. Table~\ref{tab:code_freq} shows that CoLa-ICD achieves the highest Micro-F1 across all frequency buckets on both MIMIC-III ICD-9 and MIMIC-IV ICD-10. This result indicates that the proposed framework remains effective not only on the standard ICD-9 benchmark, but also under the larger ICD-10 label space.

The improvements are particularly clear in long-tail codes. On MIMIC-IV ICD-10, CoLa-ICD improves over GKI-ICD by 0.056 and 0.077 in the 11--50 and 1--10 buckets, respectively. A similar pattern appears on MIMIC-III ICD-9, where CoLa-ICD improves the 1--10 bucket from 0.132 to 0.194. These results suggest that knowledge-enhanced evidence-label alignment helps the model distinguish rare codes.

\subsubsection{Overall Performance}
We report overall performance to assess whether the long-tail gains come at the cost of standard full-label evaluation. Table~\ref{tab:results} reports the main results on ICD-9 and ICD-10. Overall, CoLa-ICD achieves the best performance on both datasets. On ICD-9, CoLa-ICD obtains the highest Macro-F1 (0.135), Micro-F1 (0.621), P@8 (0.781), and P@15 (0.633), outperforming the strongest knowledge-enhanced baseline.

MIMIC-IV ICD-10 evaluates CoLa-ICD under a larger and sparser label space. CoLa-ICD obtains the best Macro-F1 (0.108), Micro-F1 (0.600), and P@8 (0.723), improving over the strongest baseline by 0.039, 0.021, and 0.021 absolute points, respectively. These results provide additional evidence under an ICD-10 label space. Nevertheless, the Macro-F1 remains low and should be interpreted with care, as MIMIC datasets are strongly affected by long-tailed label distributions and rare-code sparsity \citep{edin_automated_2023}.
\section{Analysis \& Discussion}

\subsection{Ablation Study}
\begin{table*}[t]
\centering
\footnotesize
\resizebox{0.85\textwidth}{!}{
\begin{tabular}{lcccccccc}
\toprule
\multirow{2}{*}{\textbf{Model}}
& \multicolumn{4}{c}{\textbf{MIMIC-III ICD-9}}
& \multicolumn{4}{c}{\textbf{MIMIC-IV ICD-10}} \\
\cmidrule(lr){2-5}\cmidrule(lr){6-9}
& \textbf{Macro-F1} & \textbf{Micro-F1} & \textbf{P@8} & \textbf{Rare Micro-F1}
& \textbf{Macro-F1} & \textbf{Micro-F1} & \textbf{P@8} & \textbf{Rare Micro-F1} \\
\midrule
CoLa-ICD
& \textbf{0.135}\textsuperscript{*} & \textbf{0.621}\textsuperscript{*} & \textbf{0.781}\textsuperscript{*} & \textbf{0.194}\textsuperscript{*}
& \textbf{0.108}\textsuperscript{*} & \textbf{0.600}\textsuperscript{*} & \textbf{0.723}\textsuperscript{*} & \textbf{0.204}\textsuperscript{*} \\
\midrule
w/o LACL
& 0.121 & 0.600 & 0.760 & 0.167
& 0.087 & 0.583 & 0.708 & 0.173 \\
w/o KI
& 0.105 & 0.588 & 0.754 & 0.153
& 0.070 & 0.569 & 0.697 & 0.155 \\
w/o co-GNN
& 0.128 & 0.602 & 0.767 & 0.174
& 0.091 & 0.589 & 0.710 & 0.181 \\
\bottomrule
\end{tabular}}
\caption{Ablation study on MIMIC-III ICD-9 and MIMIC-IV ICD-10. Rare Micro-F1 denotes Micro-F1 computed on codes in the 1--10 training-frequency bucket. ``w/o'' denotes removing one component from CoLa-ICD. KI denotes knowledge injection, LACL denotes label-aware contrastive learning, and co-GNN denotes the co-occurrence GNN. \textsuperscript{*} denotes $p<0.05$ after Holm-Bonferroni correction under the Wilcoxon signed-rank test.}
\label{tab:ablation}
\end{table*}
We conduct ablation studies on both MIMIC-III ICD-9 and MIMIC-IV ICD-10 to examine whether each component contributes consistently across different ICD label spaces. Table~\ref{tab:ablation} shows that the full CoLa-ICD model achieves the best F1 and rare-code performance on both datasets. Removing any component reduces Macro-F1, Micro-F1, P@8, and Rare Micro-F1, indicating that label knowledge, evidence-label alignment, and code-relation provide gains.

\textbf{Effect of Knowledge Injection.}
Knowledge injection makes the largest contribution to rare-code prediction. Removing it reduces Rare Micro-F1 from 0.194 to 0.153 on ICD-9 and from 0.204 to 0.155 on ICD-10
. The larger drop in ICD-10 suggests that enriched label semantics become more important as the label space grows.

\textbf{Effect of Label-Aware Contrastive Learning.}
Removing LACL consistently degrades F1 and rare-code performance on both datasets. For Rare Micro-F1, removing LACL produces drops of 0.027 on ICD-9 
and 0.031 
on ICD-10. This result supports the role of LACL in improving alignment between evidence and labels. LACL helps the model distinguish clinically similar ICD codes.

\textbf{Effect of the Co-occurrence GNN.} 
Removing the co-occurrence GNN leads to drops across the reported metrics. For Rare Micro-F1, removing it produces drops of 0.02 
on ICD-9 and 0.023 
on ICD-10. This suggests that code relation provides structural information, thereby improving prediction among correlated ICD codes.

\subsection{Evidence Utility of Attention-Selected Spans}
To investigate whether CoLa-ICD selects relevant evidence, we conduct a span removal test. If the selected span supports evidence-label alignment, removing it should result in a larger decrease in the target-code predicted probability than removing a random span. For each target code, we remove the span with the highest attention weight for that code and measure the resulting drop in the target code predicted probability. In our implementation, a span is defined as a contiguous 5 token window in the tokenized input. We select the window with the largest attention scores and measure the decrease in the probability after removing it. As a control, we remove a window of the same length as the one sampled in the document

\begin{table}[t]
\centering
\setlength{\tabcolsep}{4pt}
\renewcommand{\arraystretch}{1}
\resizebox{\linewidth}{!}{
\begin{tabular}{lcccccc}
\toprule
\textbf{Model}
& \multicolumn{3}{c}{\textbf{All codes}}
& \multicolumn{3}{c}{\textbf{Rare codes}} \\
\cmidrule(lr){2-4}\cmidrule(lr){5-7}
& \textbf{Top-1} & \textbf{Random} & \textbf{Gap}
& \textbf{Top-1} & \textbf{Random} & \textbf{Gap} \\
\midrule
CoLa-ICD & 0.097 & 0.023 & +0.074\textsuperscript{*} & 0.113 & 0.037 & +0.076\textsuperscript{*} \\
\midrule
w/o KI & 0.073 & 0.025 & +0.048\textsuperscript{*} & 0.091 & 0.031 & +0.060\textsuperscript{*} \\
w/o LACL & 0.053 & 0.024 & +0.029\textsuperscript{*} & 0.089 & 0.033 & +0.056\textsuperscript{*} \\
w/o co-GNN & 0.046 & 0.019 & +0.027\textsuperscript{*} & 0.075 & 0.029 & +0.046\textsuperscript{*} \\
\bottomrule
\end{tabular}
}
\caption{Span-removal intervention on attention-selected evidence. Values are drops in the target-code predicted probability after removing either the top-attended span (Top-1) or a random span (Random). Gap denotes Top-1 minus Random; larger gaps indicate stronger prediction-relevant evidence. \textsuperscript{*} denotes $p<0.05$ after Holm-Bonferroni correction under the Wilcoxon signed-rank test.}
\label{tab:evidence_utility}
\end{table}

Table~\ref{tab:evidence_utility} shows that attention-selected spans are more prediction-relevant than arbitrary text. For CoLa-ICD, removing the Top-1 attended span causes a larger drop in target-code predicted probability than removing a random span. This result suggests that the attention module selects spans that are helpful for the ICD decisions.

The effect is stronger for rare codes. For CoLa-ICD, removing the top-attended rare-code span lowers the target-code predicted probability by 0.113, compared with 0.097 overall and 0.037 under random rare-code span removal. This suggests that rare-code predictions rely on code-specific evidence. This aligns with the long-tail setting, where rare labels benefit from precise evidence-label alignment.

The Top-1--Random gap is also the largest for the full CoLa-ICD model. Removing KI, LACL, or the co-occurrence GNN reduces the overall gap from 0.074 to 0.048, 0.029, and 0.027, respectively. A similar pattern appears for rare codes. These results suggest that the proposed components do not merely improve classification scores but also help the model select spans with stronger predictive utility.

\subsection{Case Study}
\begin{figure}[t]      
  \centering
  \includegraphics[width=1\linewidth]{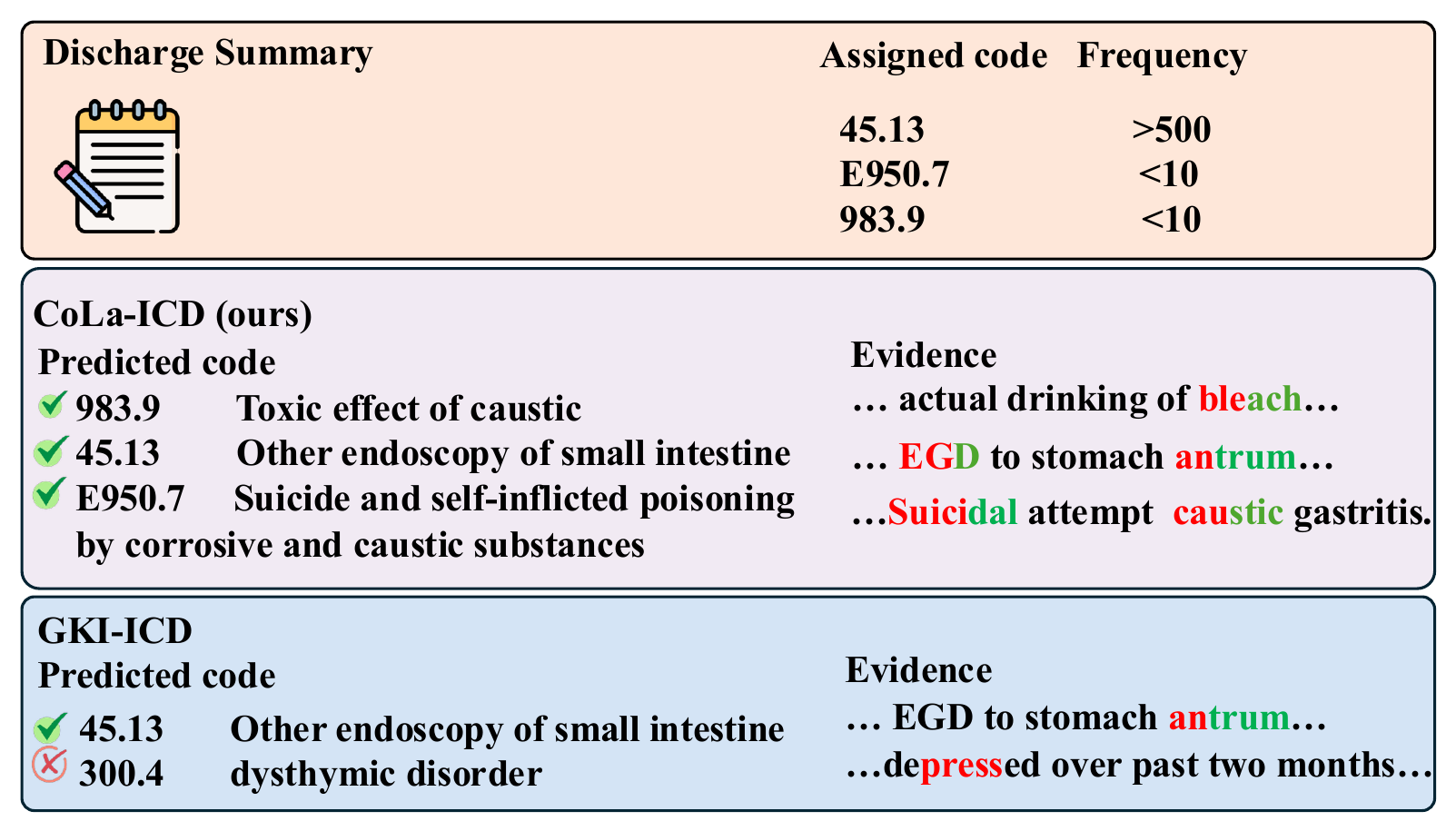}
  \caption{Case study on MIMIC-III dataset. Visualization of token-level evidence from our model and a knowledge-injection baseline (GKI-ICD). Highlighted tokens indicate those with the highest attention weights, serving as evidence.}
  \label{fig:attention}
\end{figure}
We compare CoLa-ICD with GKI-ICD, the strongest baseline in our experiments, to examine whether the performance gains are accompanied by more prediction-relevant evidence selection. The selected case includes one frequent code and two long-tail codes, allowing us to compare evidence patterns across different label-frequency regimes. For the frequent code 45.13, CoLa-ICD focuses on a broader range of clinical cues. This behavior indicates that the model aggregates multiple supportive signals. In contrast, GKI-ICD focuses on single tokens, such as "antrum", due to co-occurrence bias. However, these tokens have limited relevance to small intestine endoscopy.

For the long-tail code 983.9, CoLa-ICD identifies the key phrase "actual drinking of bleach". Our model connects it to the broader concept of caustic exposure. GKI-ICD failed to predict the correct code. For E950.7, CoLa-ICD uses the phrase “suicidal attempt” in the  context to match self-inflicted poisoning. This procedure facilitates precise predictions for both frequent and long-tail codes.

\section{Conclusion}
We introduced CoLa-ICD, a knowledge-enhanced framework for long-tail automated medical coding. CoLa-ICD enriches ICD label representations with synonyms and LLM-generated terms, models code relations with a co-occurrence GNN, and aligns clinical evidence with label semantics through multi-head term attention and label-aware contrastive learning. Experiments on ICD-9 and ICD-10 datasets show that CoLa-ICD improves overall coding performance. These findings suggest that knowledge-enhanced evidence-label alignment is a promising direction for rare-code prediction in automated medical coding.

\section{Limitations}
Although CoLa-ICD is evaluated on both ICD-9 and ICD-10, the experiments are limited to the MIMIC datasets. The MIMIC datasets are derived from a single medical center and primarily reflect ICU discharge summaries. Performance may vary across institutions, specialties, and note types. In addition, evaluation is affected by splits in prior work. Many codes appear only a few times in the training set, which makes performance sensitive to split composition. Finally, ICD coding is influenced by administrative and billing requirements, so the target labels may not always correspond perfectly to explicitly stated clinical evidence in the note.
\bibliography{references,references2}
\clearpage
\appendix

\section{Additional Hyperparameter Details}
\label{app:hyperparams}

For the label-aware contrastive learning module, the hard-negative fraction $\rho$ was selected from $\{0.1, 0.3, 0.5\}$ based on validation performance. The temperature parameter $\tau$ was set to 0.1 following previous contrastive learning work \cite{gao_simcse_2021}. The number of negative samples $K$ was also tuned on the validation set: we set $K=128$ for the main full-label experiments to cover the large label space and $K=50$ for the supplementary MIMIC-III Top50 experiment to use all available negative codes. These settings provided a balance between performance and computational cost. The global prediction threshold $\delta$ was selected on the validation set by maximizing Micro-F1 over $\delta\in[0.1,0.9]$ with a step size of 0.01.

\begin{figure}[htbp]
\centering
\includegraphics[width=\linewidth]{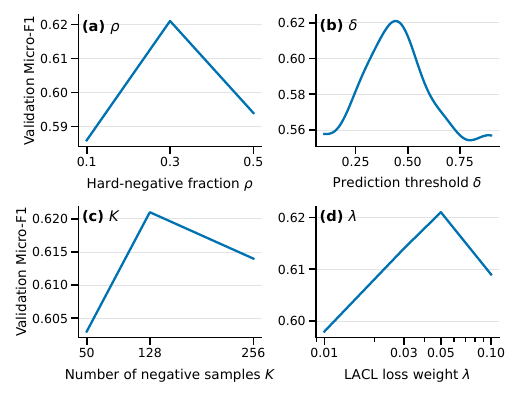}
\caption{Hyperparameter tuning on the validation set. Each panel reports validation Micro-F1 when varying one hyperparameter while keeping the others fixed: (a) hard-negative fraction $\rho$, (b) prediction threshold $\delta$, (c) number of negative samples $K$, and (d) LACL loss weight $\lambda$. Based on validation performance, we select $\rho=0.3$, $K=128$, and $\lambda=0.05$; the prediction threshold $\delta$ is selected by maximizing validation Micro-F1.}
\label{fig:hyperparameter_sensitivity}
\end{figure}

\section{Supplementary Results}
\label{app:supp_results}

\subsection{Dataset Statistics}
\label{app:dataset_stats}

\begin{table}[htbp]
\centering
\small
\begin{tabular}{lcc}
\toprule
 & \textbf{MIMIC-III ICD-9} & \textbf{MIMIC-IV ICD-10}  \\
\midrule
\#Docs &52723&209326 \\
\#Patients  &41126 &97709 \\
\#Codes &8929 &26096\\
\bottomrule
\end{tabular}
\caption{Dataset statistics of MIMIC-III(1.4) and MIMIC-IV(3.1).}
\label{tab:dataset_stats}
\end{table}

\subsection{Knowledge-Source Ablation}
\label{app:source_ablation}

To examine the contribution of different terminology sources, we compare four knowledge Injection settings: official ICD descriptions, descriptions with UMLS synonyms, descriptions with LLM-generated terms, and the full Knowledge Injection. All other training and evaluation settings are fixed. Results are averaged over ten random seeds.

Both UMLS synonyms and LLM-generated terms improve over official ICD descriptions alone across both datasets. The two terminology sources show different strengths across metrics, while their combination achieves the highest value for every reported metric. The improvement is particularly pronounced for rare codes: Full KI increases Rare Micro-F1 from 0.153 to 0.194 on MIMIC-III and from 0.155 to 0.204 on MIMIC-IV. These results suggest that curated synonyms and generated clinical expressions provide complementary lexical coverage for ICD labels.

\begin{table}[htbp]
\centering
\scriptsize
\setlength{\tabcolsep}{3pt}
\resizebox{\columnwidth}{!}{
\begin{tabular}{lcccc}
\toprule
\textbf{Knowledge Source}
& \textbf{Macro-F1}
& \textbf{Micro-F1}
& \textbf{P@8}
& \textbf{Rare Micro-F1} \\
\midrule
\multicolumn{5}{c}{\textbf{MIMIC-III ICD-9}} \\
\midrule
Description      & 0.105 & 0.588 & 0.754 & 0.153 \\
+ UMLS synonyms  & 0.118 & 0.605 & 0.759 & 0.168 \\
+ LLM terms      & 0.113 & 0.608 & 0.767 & 0.173 \\
\textbf{Full KI} & \textbf{0.135} & \textbf{0.621} & \textbf{0.781} & \textbf{0.194} \\
\midrule
\multicolumn{5}{c}{\textbf{MIMIC-IV ICD-10}} \\
\midrule
Description      & 0.070 & 0.569 & 0.697 & 0.155 \\
+ UMLS synonyms  & 0.086 & 0.581 & 0.701 & 0.171 \\
+ LLM terms      & 0.083 & 0.586 & 0.703 & 0.173 \\
\textbf{Full KI} & \textbf{0.108} & \textbf{0.600} & \textbf{0.723} & \textbf{0.204} \\
\bottomrule
\end{tabular}
}
\caption{Knowledge-source ablation on MIMIC-III ICD-9 and MIMIC-IV ICD-10.}
\label{tab:source_ablation}
\end{table}
\FloatBarrier

\subsection{MIMIC-III Top50 Results}
\label{app:top50}
We report MIMIC-III Top50 results for comparability with prior automated medical coding work. This subset focuses on frequent labels and is therefore not used as the main evidence for long-tail ICD coding.

\begin{table}[htbp]
\centering
\small
\resizebox{\linewidth}{!}{
\begin{tabular}{lccccc}
\toprule
\multirow{2}{*}{\textbf{Model}} & \multicolumn{2}{c}{\textbf{AUC}} & \multicolumn{2}{c}{\textbf{F1}} & \multirow{2}{*}{\textbf{P@5}} \\
\cmidrule(lr){2-3}\cmidrule(lr){4-5}
 & Macro & Micro & Macro & Micro & \\
\midrule
CAML & 0.875 & 0.909 & 0.532 & 0.614 & 0.609 \\
MSMN & 0.928 & 0.947 & 0.683 & 0.725 & 0.680 \\
KEPTLongformer & 0.926 & 0.947 & 0.689 & 0.728 & 0.672 \\
PLM-ICD & 0.910 & 0.934 & 0.663 & 0.719 & 0.660 \\
PLM-CA & 0.916 & 0.936 & 0.671 & 0.710 & 0.664 \\
CoRelation & \textbf{0.933} & 0.951 & 0.693 & 0.731 & \underline{0.683} \\
GKI-ICD & \textbf{0.933} & \underline{0.952} & 0.692 & \underline{0.735} & 0.681 \\
\textbf{CoLa-ICD (Ours)} & \underline{0.932} & \textbf{0.956}\textsuperscript{*} & \textbf{0.698}\textsuperscript{*} & \textbf{0.741}\textsuperscript{*} & \textbf{0.712}\textsuperscript{*} \\
\bottomrule
\end{tabular}
}
\caption{Supplementary results on MIMIC-III Top50. Bold indicates the best result, underlining indicates the second best, and \textsuperscript{*} denotes $p<0.05$ after Holm-Bonferroni correction under the Wilcoxon signed-rank test.}
\label{tab:top50_appendix}
\end{table}

\subsection{MIMIC-IV ICD-9 Results}
\label{app:mimiciv_icd9}
We report MIMIC-IV ICD-9 results as supplementary evidence for comparability with prior work. These results are not used as the main ICD-10 generalization evidence in the paper.

\begin{table}[htbp]
\centering
\small
\resizebox{\linewidth}{!}{
\begin{tabular}{lccccc}
\toprule
\multirow{2}{*}{\textbf{Model}} & \multicolumn{2}{c}{\textbf{AUC}} & \multicolumn{2}{c}{\textbf{F1}} & \multirow{2}{*}{P@8}\\
\cmidrule(lr){2-3} \cmidrule(lr){4-5}
 & Macro & Micro & Macro & Micro  \\
\midrule
MSMN      & 0.972 & 0.993 & 0.277 & 0.618 & 0.691 \\
PLM-CA   & 0.961 & 0.992 & 0.184 & 0.589& 0.666 \\
GKI-ICD &\underline{0.973}&\underline{0.994}&\underline{0.298}&\underline{0.623}&\underline{0.703} \\
\textbf{CoLa-ICD (Ours)}   & \textbf{0.978}\textsuperscript{*}& \textbf{0.996}\textsuperscript{*}& \textbf{0.337}\textsuperscript{*}& \textbf{0.639}\textsuperscript{*}& \textbf{0.720}\textsuperscript{*} \\
\bottomrule
\end{tabular}
}
\caption{Supplementary results on MIMIC-IV ICD-9. Bold indicates the best result, underlining indicates the second best, and \textsuperscript{*} denotes $p<0.05$ after Holm-Bonferroni correction under the Wilcoxon signed-rank test.}
\label{tab:mimiciv_icd9_appendix}
\end{table}

\section{Prompts for Term Generation}
\label{prompts for term}
We used the Gemini 2.5 Flash API to generate abbreviations, common expressions,
and extended descriptions for ICD codes.
Below we provide an example prompt template 
(the placeholders \texttt{\{icd\_code\}} and \texttt{\{icd\_desc\}} 
are replaced with the actual code and description at generation).

\begin{promptbox}{Prompt Template for ICD Term Generation}
You are a senior expert in medical terms and clinical documentation with extensive knowledge of ICD-9 or ICD-10 coding systems.
Given the ICD code \texttt{\{icd\_code\}} with the description \texttt{\{icd\_desc\}},

\textbf{Task:}
For the ICD code below, generate three lists:

1) \texttt{"abbr"}: commonly used abbreviations for this diagnosis as they appear in clinical notes.

2) \texttt{"common"}: short, colloquial clinical expressions for the same diagnosis as typically written in EHR text.

3) \texttt{"extend"}: extended definitions that provide more descriptive context.

\textbf{Guidelines}

1. \textbf{Medical Accuracy:} Maintain precise medical terms and accurate ICD-9 terms, standard abbreviations, and peer-reviewed nomenclature.

2. \textbf{Scope:} Do not add other diagnoses, etiologies, stages, or qualifiers not implied by the original description. Use different sentence structures and medical phrasing approaches.

3. \textbf{Ambiguity:} Omit highly ambiguous abbreviations that are not strongly associated with this diagnosis across routine practice.
\end{promptbox}

\section{LLM-Generated Term Selection}
\label{app:llm_terms_selection}
This section provides additional details on how we evaluate the five LLM-generated candidate term sets and select the final terms for each ICD code. The procedure is carried out \textbf{only} on the MIMIC-III training and validation splits. The selected strategy is applied unchanged to MIMIC-IV. These implementation notes complement the brief description in Section \ref{sec_know_inject} and are included here for reproducibility.
\subsection{Classifier architecture and inputs}
We use the BioClinicalBERT encoder followed by a trainable two-layer MLP classification head. Given a discharge summary $X$, we obtain the text representations
$H \in \mathbb{R}^{T \times d}$. For each ICD code $c_i$ and one candidate term set $T_i^{(r)}$ ($r \in \{1,\dots,5\}$), we encode all terms in $T_i^{(r)}$ with BioClinicalBERT and feed them into the classifier to get the result. BioClinicalBERT is kept frozen and only the term-attention module and linear classifier are trainable.
\subsection{Training setup and objective.}
For each candidate index $r$, we instantiate one classifier that uses $\{T_i^{(r)}\}_{i=1}^N$ as its term set. Each classifier is trained independently on the MIMIC-III training split with binary cross-entropy loss over all $N$ labels, for 10 epochs, with batch size 8 and learning rate $2\times10^{-5}$. We tune and select terms \emph{only} on the official validation split.

\subsection{Term Set Scoring and Selection}
To determine the optimal term source, we evaluate the five candidate term sets on the validation split. Let $\mathcal{S}_r$ denote the $r$-th candidate set, which contains descriptions for all ICD codes generated by a specific prompt strategy.
We select the best candidate set index $r^{\star}$ based on the overall Macro-F1 score:
\begin{equation}
    r^{\star} = \operatorname*{arg\,max}_{r \in \{1,\dots,5\}} \text{Macro-F1}(\mathcal{D}_{val}; \mathcal{S}_r).
\end{equation}
Once the optimal source $r^{\star}$ is identified, we use it to train the final CoLa-ICD model.

\end{document}